\documentclass[10pt,letterpaper]{article}
\usepackage[marginparwidth=2in]{geometry}

\usepackage[utf8]{inputenc}

\usepackage{cite}
\usepackage{float}
\usepackage{nameref,hyperref}

\usepackage{microtype}
\DisableLigatures[f]{encoding = *, family = * }

\usepackage{changepage}

\usepackage[aboveskip=1pt,labelfont=bf,labelsep=period,singlelinecheck=off]{caption}

\makeatletter
\renewcommand{\@biblabel}[1]{\quad#1.}
\makeatother

\usepackage{lastpage,fancyhdr,graphicx}
\usepackage{epstopdf}
\usepackage{color}
\usepackage{footmisc}
\definecolor{Gray}{gray}{.25}

\usepackage{graphicx}

\usepackage{sidecap}

\usepackage{wrapfig}
\usepackage[pscoord]{eso-pic}
\usepackage[fulladjust]{marginnote}
\reversemarginpar
\begin{document}
	
	\begin{center}
		\textbf{\Large
			Fully Convolutional Networks for Automated Segmentation of Abdominal Adipose Tissue Depots in Multicenter Water-Fat MRI}\\
		\vspace{0.5cm}
		\begin{large}
			Taro Langner\textsuperscript{1*},
			Anders Hedström\textsuperscript{2},			
			Katharina Paulmichl\textsuperscript{3,4},
			Daniel Weghuber\textsuperscript{3,4},\\
			Anders Forslund\textsuperscript{5},
			Peter Bergsten\textsuperscript{5,6},
			Håkan Ahlström\textsuperscript{1,2},
			Joel Kullberg\textsuperscript{1,2}
		\end{large}
		\\
		\bigskip
	\end{center}

	\textsuperscript{1}Dept. of Radiology, Uppsala University, Uppsala, Sweden
	\\
	\textsuperscript{2}Antaros Medical, BioVenture Hub, Mölndal, Sweden
	\\
	\textsuperscript{3}Dept. of Pediatrics, Paracelsus Medical University, 5020 Salzburg, Austria
	\\
	\textsuperscript{4}Obesity Research Unit, Paracelsus Medical University, 5020 Salzburg, Austria
	\\
	\textsuperscript{5}Dept. of Women’s and Children’s Health, Uppsala University, Uppsala, SE 751 05, Sweden
	\\
	\textsuperscript{6}Dept. of Medical Cell Biology, Uppsala University, Uppsala, SE 751 23, Sweden
	\\

	\bigskip
	*taro.langner@surgsci.uu.se

	\bigskip
	\textbf{Published in \textit{Magnetic Resonance in Medicine}}
	\bigskip
	\section*{Abstract}
	\textbf{Purpose:} An approach for the automated segmentation of visceral adipose tissue (VAT) and subcutaneous adipose tissue (SAT) in multicenter water-fat MRI scans of the abdomen was investigated, using two different neural network architectures.\\
	\textbf{Methods:} The two fully convolutional network architectures U-Net and V-Net were trained, evaluated and compared on the water-fat MRI data. Data of the study Tellus with 90 scans from a single center was used for a 10-fold cross-validation in which the most successful configuration for both networks was determined. These configurations were then tested on 20 scans of the multicenter study beta-cell function in JUvenile Diabetes and Obesity (BetaJudo), which involved a different study population and scanning device.\\
	\textbf{Results:} The U-Net outperformed the used implementation of the V-Net in both cross-validation and testing. In cross-validation, the U-Net reached average dice scores of $0.988$ (VAT) and $0.992$ (SAT). The average of the absolute quantification errors amount to $0.67\%$ (VAT) and $0.39\%$ (SAT). On the multi-center test data, the U-Net performs only slightly worse, with average dice scores of $0.970$ (VAT) and $0.987$ (SAT) and quantification errors of $2.80\%$ (VAT) and $1.65\%$ (SAT).\\
	\textbf{Conclusion:} The segmentations generated by the U-Net allow for reliable quantification and could therefore be viable for high-quality automated measurements of VAT and SAT in large-scale studies with minimal need for human intervention. The high performance on the multicenter test data furthermore shows the robustness of this approach for data of different patient demographics and imaging centers, as long as a consistent imaging protocol is used.
	
	\bigskip
	
	\textbf{Key words:} deep learning, fully convolutional networks, segmentation, water-fat MRI, adipose tissue, abdominal 
	
	
	\section*{Introduction}
	The quantification of human adipose tissue depots has the potential to provide new insights into the role of body composition as a factor for metabolic and cardiovascular disease. Abdominal obesity has been linked to conditions such as hypertension, inflammation and type 2 diabetes and is increasingly prevalent even among young adults and children \cite{despres2006abdominal}. Due to their different roles in the human metabolism, the total amount of abdominal adipose tissue is commonly separated into the two main components of visceral adipose tissue (VAT) and subcutaneous adipose tissue (SAT), with the former being more closely associated with health risks \cite{hu2016segmentation}. Other depots such as intramuscular adipose tissue and areas surrounding the spine are typically excluded. In practice the segmentation of VAT and SAT is usually performed with the help of automatic or semi-automatic methods which often require manual inspection and corrections by human experts. In larger studies with dozens or hundreds of scanned volumes this results in a high workload, so that an accurate, automated strategy that minimizes the need for human input and yields consistent results has the potential to reduce the cost and increase the feasibility of large scale studies.\\
	
	For measurements of the quantity and distribution of adipose tissue in medical research, non-invasive imaging methods are commonly employed such as CT and MRI. When using chemical-shift encoded water-fat MRI, it is possible to obtain both co-registered water and fat signal images as well as voxel-wise fat fraction values \cite{hu2013quantitative} without exposing the patient to ionizing radiation.

	A variety of automated and semi-automated methods for VAT and SAT segmentation have been developed for images derived from both CT and MRI. In CT images, techniques have been proposed that use thresholding followed by ray tracing to isolate areas that are surrounded by lean tissue \cite{kullberg2017automated}. However, methods like this depend on the use of the Hounsfield scale, so that they can not be directly applied to MR images.
	In whole-body MRI of mice, a previously presented method used a combination of clustering and competitive region growing for the identification of narrow passages which delineate the depots of VAT and SAT \cite{ranefall2009automatic}. To simplify the task of segmentation, a common strategy consists in imaging and segmenting only a single transverse slice located between the vertebrae L2-L5 as an indicator for overall VAT and SAT. However, it has been noted that the the accuracy of this approach is insufficient \cite{addeman2015validation}, \cite{shen2016automatic}, so that a volumetric assessment is expected to result in more reliable measurements. In water-fat MRI, further strategies have been proposed such as the transfer of segmentations between volumes \cite{joshi2013automatic}, clustering techniques for masking and fitting of three-dimensional surfaces \cite{addeman2015validation} or morphological operators that allow the identification of VAT and SAT. \\
	
	Rather than relying on techniques such as clustering, thresholding and registration, the most successful methods for image-based semantic segmentation on current benchmark datasets in the computer vision community employ machine learning strategies such as convolutional neural networks \cite{everingham2015pascal}, \cite{garcia2017review}, \cite{lecun2015deep}, which have also seen success in medical applications \cite{shen2017deep}. For segmentation tasks, a network trained for classification can be applied in a sliding window technique to patches of an image to predict a label for each given central voxel. This approach has been previously used to generate adipose tissue segmentations in CT images \cite{wang2017two}. However, the redundant feature extraction for adjacent patches by the sliding window technique is highly inefficient, so that specialized architectures for segmentation have emerged.

	Based on the concept of fully convolutional networks \cite{long2015fully}, derived architectures have been proposed such as the U-Net \cite{ronneberger2015u} for the segmentation of two-dimensional biomedical cell images and the V-Net \cite{milletari2016v} for segmentation of the human prostate in three-dimensional MR images. 
	We therefore introduce a new approach using a fully convolutional network for the automated segmentation of VAT and SAT, with the goal of investigating how high of an accuracy and robustness can be achieved on this task. The network was applied to a representation of the water-fat MRI scans in which the water and fat signal as well as the calculated voxel-wise fat fractions are combined. Both the U-Net and the V-Net were adapted for this task and their performance was compared in both a ten-fold cross-validation as well as on a separate test dataset containing images from two different centers, each using a different MR system.

	\section*{Methods}
	
	The water-fat MRI data obtained from two separate studies was used to train and evaluate the performance of convolutional neural network architectures for semantic segmentation. The image data of the study Tellus was used for the training process, in which the learnable parameters of the network are adjusted, as well as the validation phase, in which a chosen network configuration together with its learned parameters is evaluated. The most successful network configurations were then tested on images of the study beta-cell function in JUvenile Diabetes and Obesity (BetaJudo) \cite{staaf2017pancreatic}. \\
	
	\subsubsection*{Water-fat MRI Data}
	
	The image data of the study Tellus was acquired from a cohort of adult male and female subjects, diagnosed with type 2 diabetes, aged 18-80 years, with a body mass index of up to $40kg/m^2$. The images were acquired with a 1.5T MR system Achieva dStream, Philips Healthcare, Best, The Netherlands at the University Hospital Uppsala, Sweden with anterior and posterior coils using the mDixon Quant sequence with TR 5.6ms, TE1 0.95ms, deltaTE 0.7ms, flip angle 5deg and fixed FOV. Among these patients, 45 were selected who participated in two visits, between 27 and 45 days apart, yielding 90 scan volumes with a typical image resolution of [256, 256, 20] voxels of size [2.07, 2.07, 8] mm. 
	
	The image data of the study BetaJudo \cite{staaf2017pancreatic} was acquired in a collaboration between the Paracelsus Medical University Hospital in Salzburg, Austria and Uppsala University Hospital, Sweden. The study population consisted of 116 male and female individuals between the age of 10 and 18 years with complete records and includes both normal-weight and overweight subjects. A standardized imaging protocol was applied, with the Uppsala center using the same scanner and configuration as in the Tellus study, and the Salzburg center using a 1.5T Philips Ingenia system with TR 8.8ms, TE1 1.38ms, deltaTE 2.6ms, flip angle 5deg, again with fixed FOV. The typical image resolution for these scans is [256, 176, 21] voxels, again of size [2.07, 2.07, 8] mm. From this population 10 subjects were randomly selected for each center, resulting in a total of 20 scans that formed the test data set. \\
	
	There were systematic differences between the scans of both studies. The images obtained from Tellus included the arms of the patients and were masked to contain a value of zero in the image background. In BetaJudo, the arms are not included and signal noise in the background leads to noisy fat fraction values. The volumes consist of 21 instead of 20 transverse slices and the scanned area is slightly shifted from (L4-L5) to (L3-L4), so that it includes less of the hip bone. Despite the lower age of the subjects, the labeled volumes for SAT in the chosen scans are on average about 75\% larger, while the VAT volumes are about 50\% smaller than in the images of Tellus. 
	In order to obtain reference segmentations, the images of both studies were labeled with manual input. For the Tellus data, a first estimate at the correct labels was generated using the inside lean tissue filter (4) and exclusion of voxels with fat fraction values below 50\%. The results were then manually corrected by an experienced operator by adjusting the delineation of VAT and SAT in the software SmartPaint (18) based on the water image and removing adipose tissue around the spinal column. On BetaJudo the reference segmentations were generated by another operator in a fully manual procedure using the software 3DSlicer on the fat-fraction image, likewise excluding voxels with fat fractions of less than 50\% as well as adipose tissue around the spine.
	
	\begin{figure}[t] 
		
		
		\includegraphics[width=\textwidth]{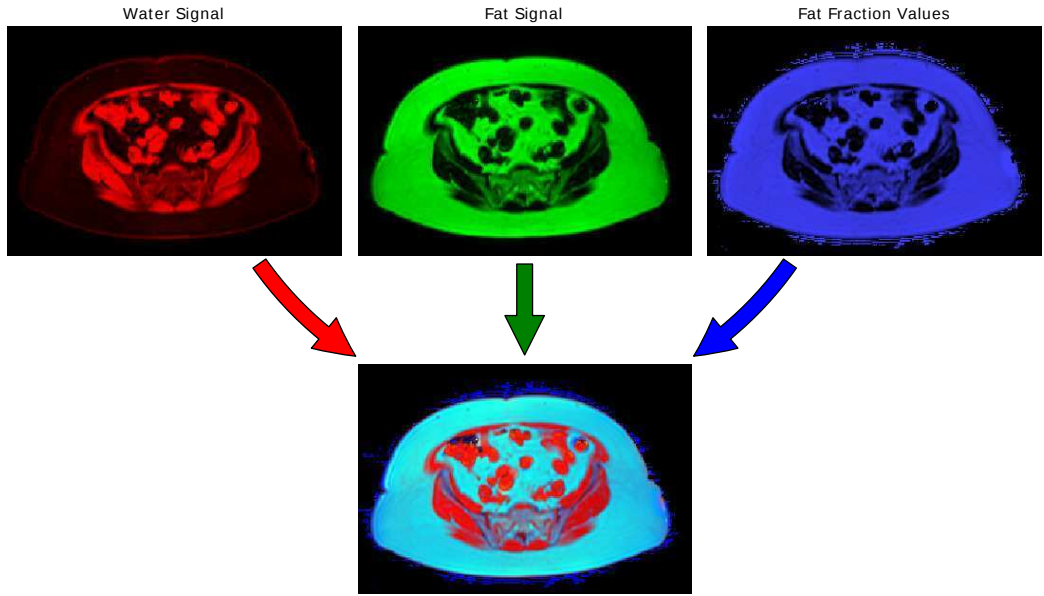}
		
		\caption{By combining the fat and water signal as well as the voxel-wise fat fraction values obtained from water-fat MRI, a three-channel image can be formed as input for the networks.}
		
		\label{fig1} 
		
	\end{figure}
	
	\subsubsection*{Data formatting}
	Several pre-processing steps were applied to the image data. In the Tellus dataset, the arms of the patients were removed from all scans. The contrast of the signal images of the water-fat MRI was then adjusted for each transverse slice by clipping the brightest one percent of their respective histograms and normalizing the remaining intensities. This strategy greatly decreases the variation in intensities seen both across different volumes as well as between the central and outer slices of a given scan where signal loss occurs. The fat fraction values were not normalized, so that the actual percentage is retained. The water and fat signal images as well as the fat fraction image were combined to form the three channels of a color image as seen in Figure Several pre-processing steps were applied to the image data. For BetaJudo, all image slices were zero-padded to a size of [256, 256] and the noise in the background was masked out. In the Tellus dataset, the arms of the patients were removed from all scans to simplify the problem under the assumption that this step could be automatically performed in the future, possibly with a conventional algorithm. The contrast of the signal images of the water-fat MRI was then adjusted for each transverse slice by clipping the brightest one percent of their respective histograms and normalizing the remaining intensities to a float value range of [0, 1] for processing by the networks. This strategy greatly decreases the variation in intensities seen both across different volumes as well as between the central and outer slices of a given scan where signal loss occurs. The fat fraction values were not normalized, so that the actual percentage is retained. The water and fat signal images as well as the fat fraction image were combined to form the three channels of a color image as seen in Figure \ref{fig1}. \\

	\subsubsection*{Automated Segmentation}
	In order to train and validate the different network architectures, the data of the Tellus study was split on the patient level for a 10-fold cross-validation. In this way, the available data was split into ten subsets, each of which was used to evaluate the performance of a network instance trained on the remaining nine sets. The presented metrics for the cross-validation are obtained by uniting the results on the individual sets. The highest-scoring network configurations were then trained once more on all available scans of Tellus in order to be tested on the data of the BetaJudo study.	
	
	These splits of the image data were used for the training of convolutional neural networks for semantic segmentation. Based on the given reference segmentations, these architectures are able to successively apply convolutional filters for the extraction of hierarchical image features that allow for a segmentation to be automatically generated. The relevant features and their role in deciding on the shape of the segmentation are learned from the reference data by supervised learning.
	Both architectures follow an encoder-decoder structure in which the representations of the input are first downsampled and later upsampled in multiple steps, with the goal of extracting both fine as well as large-scale features. Long skip connections allow for both of these types of features to be combined in order to eventually assign a label to each voxel of the input image.
	
		\begin{figure}[t] 
			
			
			\includegraphics[width=\textwidth]{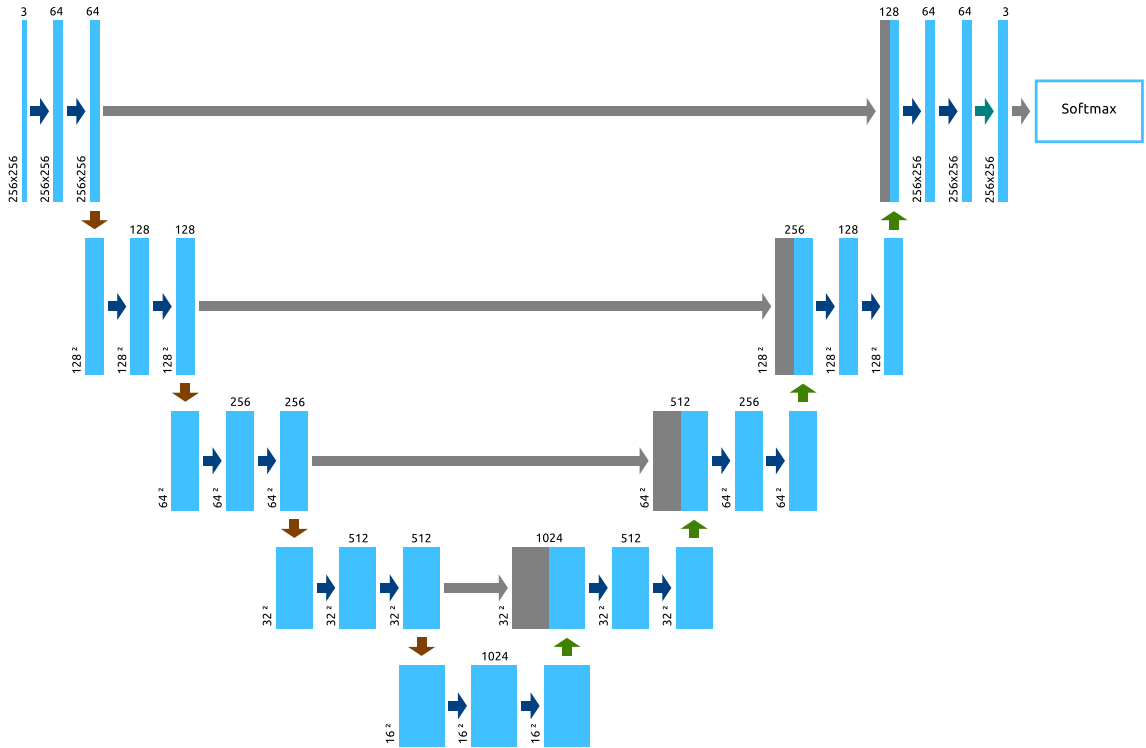}
			
			\caption{U-Net architecture used for the experiments. A two-dimensional input slice with three channels is passed to the network, yielding pixel-wise scores for all three classes. In contrast to the original architecture, zero-padding ensures that the size of the feature maps stays consistent on each level, so that no cropping is needed in the skip connections. More detail is found in the uploaded PyTorch implementation.}
			
			\label{fig2} 
			
		\end{figure}
		
			\begin{figure}[t] 
				
				
				\includegraphics[width=\textwidth]{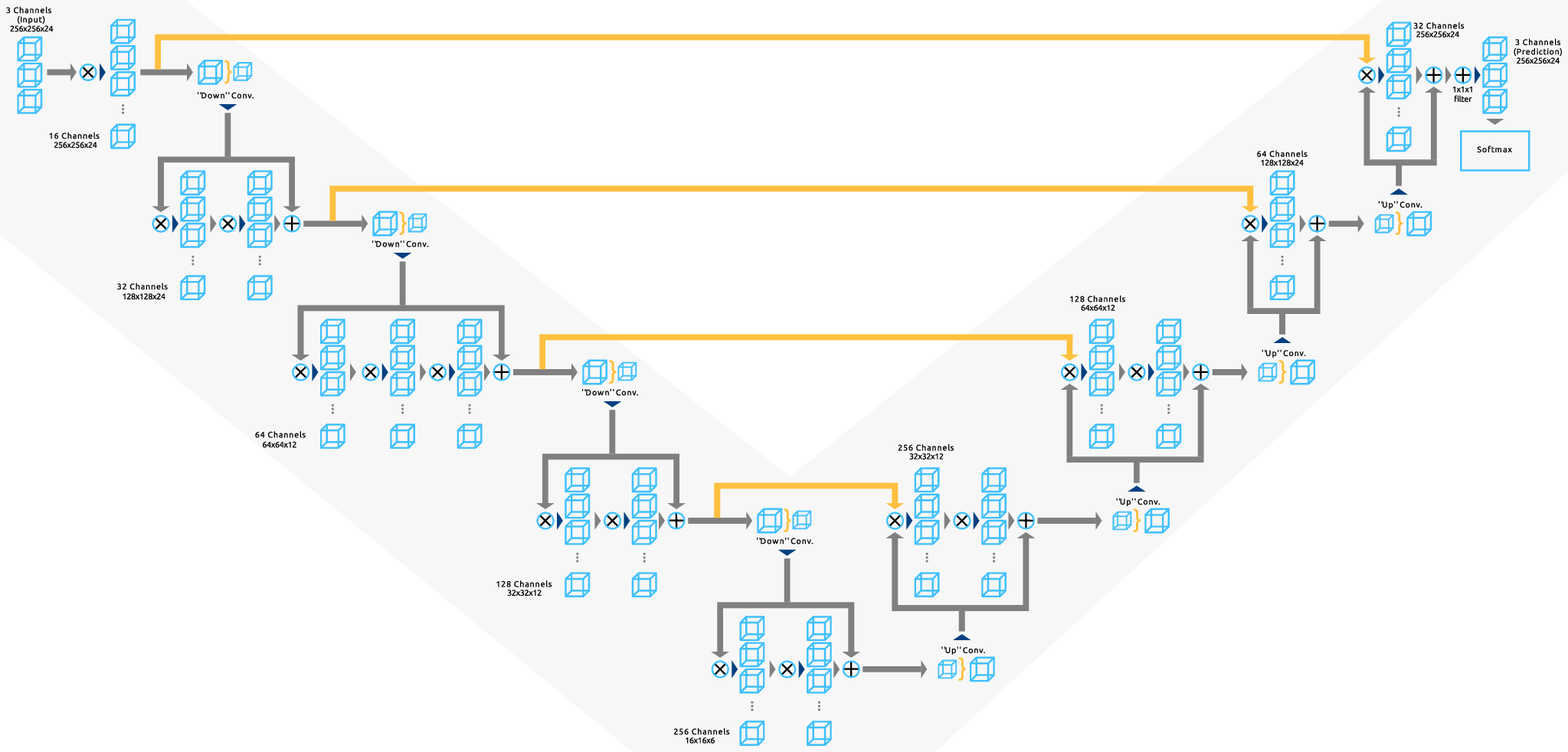}
				
				\caption{V-Net architecture used for the experiments. A three-dimensional input volume with three channels is passed to the network, yielding voxel-wise scores for all three classes. In contrast to the original architecture, no short skip connection is used in the first convolutional block. The number of convolutional steps for the lower-resolution feature maps is increased less, especially in the decoding part of the network in order to save processing time and memory requirements. Rather than simply halving the resolution of the entire volume on each level, the strided convolution is adjusted to retain the number of slices after the first and third level despite compressing the other two dimensions. More detail is found in the uploaded PyTorch implementation.}
				
				\label{fig3} 
				
			\end{figure}
	\subsubsection*{U-Net}
	As a first network architecture, the U-Net \cite{ronneberger2015u} was trained on the two-dimensional transverse slices of the chosen volumes. Preliminary experiments showed that in the chosen configuration the dice scores of the segmentations did not improve when the number of pooling layers was increased or diminished. The architecture of the original paper with four downsampling steps was therefore retained and is shown in Figure \ref{fig2}.
	Due to their relatively small size, all slices can easily be processed as a whole without any need for tiling. Furthermore internal zero padding for the filters was used instead of mirrored padding of the input image, so that input and output of the network have the same dimensions. In this way the runtime and memory requirements for the network were roughly halved and no penalty on segmentation quality was observed. This strategy has been previously reported by as successful by multiple papers \cite{dong2017automatic}, \cite{kayalibay2017cnn}.

	\subsubsection*{V-Net}
	The implementation of the V-Net \cite{milletari2016v} that was used for the following experiments is based on a GitHub repository\footnote{\url{https://github.com/mattmacy/vnet.pytorch}} with that includes modifications such as batch normalization and dropout. It is worth noting that the description of the V-Net architecture in the original paper also differs slightly from the actual implementation that was used by its authors to generate their reported results\footnote{\url{https://github.com/faustomilletari/VNet/issues/9}}. In order to process the highly anisotropic multi-channel data, the following additional architectural adjustments were necessary.

	The first adjustment affects the dimensionality reduction and is necessary due to the low number of slices. When applying a strided convolution for downsampling, the resolution of the volumetric feature maps in the V-Net is effectively halved in each step. When the third dimension of the input volumes has an extent of just 20, only the first two halving steps result in an even number of slices. Due to this restriction, all input volumes were padded to 24 slices by concatenating copies of the last slice. Additionally, the stride of the first and third strided convolution was set to a value of one along the longitudinal axis, so they do not affect the number of slices. At the end of the downsampling path the volume is thereby represented by feature maps with 6 slices. A visualization of the resulting architecture is listed in the Appendix. When evaluating the dice score and other performance metrics on the segmentation results, the padding slices were excluded so that a direct comparison to the U-Net is possible. The second change is required due to the usage of multiple image channels. In the V-Net the result of the first convolution step is combined as an element-wise sum with the original input volume by a short skip connection. This skip connection was removed to avoid a conflict between the 3 input image channels and the 16 volumetric feature map channels. Furthermore, the number of convolution steps was reduced especially in the decoder part of the architecture in order to improve the network speed. The resulting architecture is shown in Figure \ref{fig3}.

	\subsubsection*{Settings and Evaluation}
	All reported results were achieved in the framework PyTorch using the Adam optimizer \cite{kingma2014adam} and a learning rate of 0.0001 as well as a batch size of one. The learned weights were initialized with the PyTorch default settings, randomly sampled from a scaled, zero-centered Gaussian distribution1. No benefit was observed when using class weights for the loss function, so that for the reported results all classes were weighted evenly. The U-Net was trained with a pixel-wise cross-entropy loss. \\

		\begin{figure}[H] 
			
			
			\includegraphics[width=\textwidth]{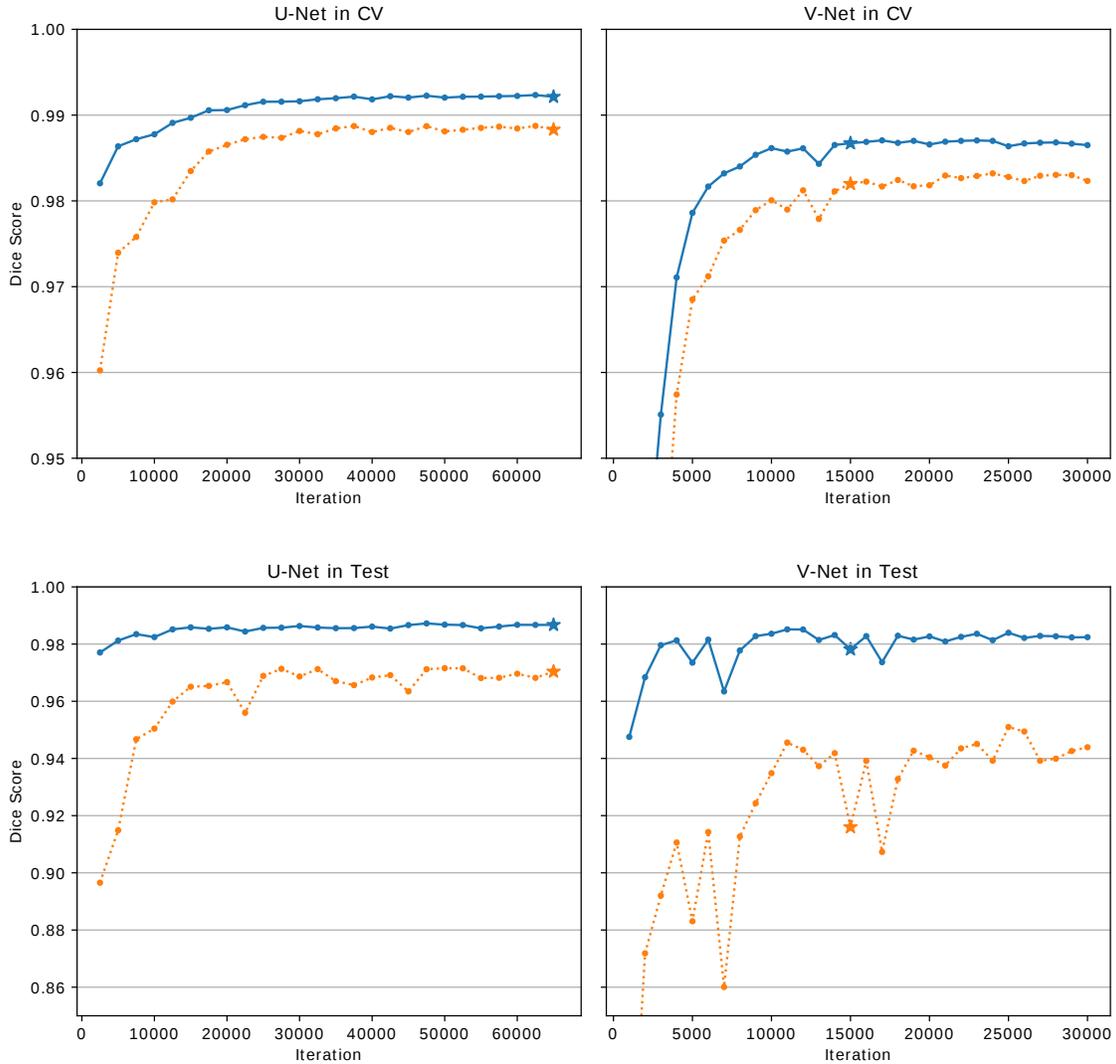}
			
			\caption{Training curves for the U-Net and V-Net in both cross-validation and on the test data. The x-axis marks the given training iteration, with the y-position representing the average dice score on the validation or test data for VAT (dotted line) and SAT (continuous line). Based on the cross-validation, the iteration marked with a star symbol was chosen to calculate the listed results.}
			
			\label{fig4} 
			
		\end{figure}

		For the V-Net an improvement over the voxel-wise cross-entropy loss was found when using a dice loss with a term alpha = $0.1$. to ensure numerical stability and avoid division by zero. For a set of voxels X labeled by the method to be evaluated and the set of voxels Y that are part of the reference segmentation, the loss function L to be minimized therefore takes the following form:

		\begin{equation}L(X,Y)=1 - \frac{2 \mid X \cap Y \mid + alpha}{\mid X\mid +\mid Y \mid + alpha}
		\end{equation}
		\label{eq1}
			
	The U-Net was trained on a Nvidia GeForce GTX 1080Ti graphics card for 65000 iterations, requiring about one hour on each training split. The V-Net, in contrast, was trained for only 15000 iterations but required about ten hours per split. The resulting training curves are seen in Figure \ref{fig4} and an implementation for the networks used in these experiments has been uploaded to GitHub\footnote{\url{https://github.com/tarolangner/fcn_vatsat}}. The results were evaluated by dice score as well as the volume and percentage of the error of the segmentation. The effect of the pre-processing steps was analyzed and additionally, systematic errors and patterns in the network performance were visually evaluated. In order to judge in how far the retrieved dice scores might be affected by differences Visualizations of both network architectures and training curves are listed in the segmentation styles of the two operators, an additional inter-operator dice score was calculated. This score is based on the segmentation of 20 randomly chosen volumes as a representative sample from the Tellus dataset, created by the operator who segmented the images of BetaJudo.

	\section*{Results}

	\begin{table}[H]
		\centering
		\caption{Average performance of the networks in cross-validation (CV) on the data of Tellus and on the test data of BetaJudo (Test). The listed error in \% is the average of all absolute differences in measured and reference volumes divided by the reference volume.}
		\vspace{0.5cm}
		\label{table_unet}
		\begin{tabular}{llrr}
			\hline
			& & \textbf{U-Net} &  \\
			Depot & Metric & CV & Test \\
			\hline
			& Dice & $0.988 \pm 0.007$ & $0.970 \pm 0.010$  \\
			VAT	& Error in \% & $0.67 \pm 0.80$ & $2.80 \pm 1.55$  \\
			& Error in mL & $1 \pm 27$ & $-41 \pm 32$ \\
			\hline
			& Dice & $0.992 \pm 0.003$ & $0.987 \pm 0.004$  \\
			SAT	& Error in \% & $0.39 \pm 0.35$ & $1.65 \pm 0.80$  \\
			& Error in mL & $2 \pm 26$ & $-112 \pm 45$ \\
			\hline
		\end{tabular}
	\end{table}	
	
	\begin{table}[H]
		\centering
		
		\label{table_vnet}
		\begin{tabular}{llrr}
			\hline
			& & \textbf{V-Net} &  \\
			Depot & Metric & CV & Test \\
			\hline
			& Dice & $0.982 \pm 0.009$ & $0.916 \pm 0.059$  \\
			VAT	& Error in \% & $1.15 \pm 1.06$ & $8.86 \pm 10.15$ \\
			& Error in mL & $-18 \pm 54$ & $75 \pm 165$ \\
			\hline
			& Dice & $0.987 \pm 0.004$ & $0.978 \pm 0.012$ \\
			SAT	& Error in \% & $0.86 \pm 0.84$ & $3.02 \pm 2.28$  \\
			& Error in mL & $24 \pm 54$ & $-231 \pm 187$  \\
			\hline
		\end{tabular}
	\end{table}	
		
	For both cross-validation and testing, the average performance metrics are seen in Table \ref{table_unet}. The U-Net consistently outperforms the V-Net and generates segmentations that have an average error of less than one percent of the volume in cross-validation. Both networks achieve a slightly lower performance on the test data, but the U-Net still reaches an average error of less than 3\% for the more difficult VAT volume. As an additional test, both networks were also evaluated on the test data without previous removal of the background noise. Although no such noise is present in the training data, the performance of the U-Net is virtually unaffected, with no change in the metrics at the significance given in Table \ref{table_unet}. The V-Net is less robust and erratically oversegments random patterns in the background noise, so that no separate metrics are listed for this test.

	\begin{figure}[H] 
		\includegraphics[width=\textwidth]{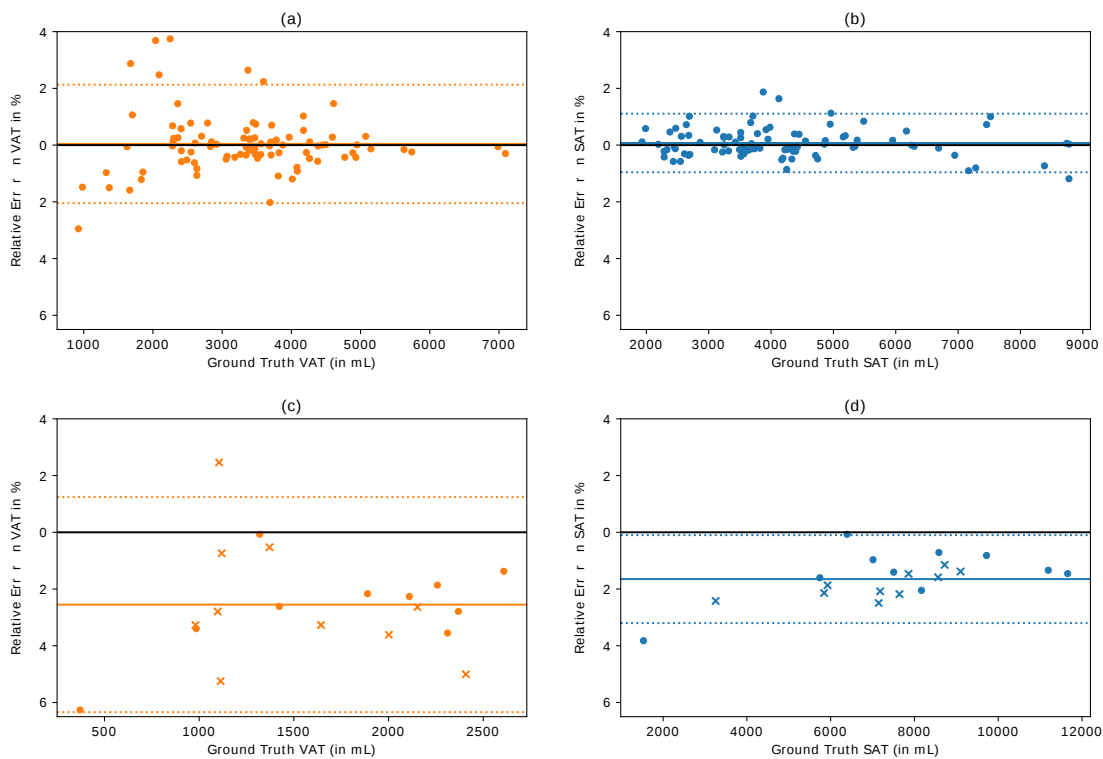}
		
		\caption{Relative error in volume measurement (y-axis) as calculated in (predicted volume / reference volume - 1) in relation to the reference volumes (x-axis) for the U-Net in cross-validation (a, b) and on the test set (c, d). The dotted lines denote two standard deviations away from the mean. Note how on the test set the data from Uppsala (dot markers) is on average closer to an error of zero than the data from Salzburg (x markers).}
		
		\label{fig5} 
		
	\end{figure}
\begin{figure}[H] 
		
		
		\includegraphics[width=\textwidth]{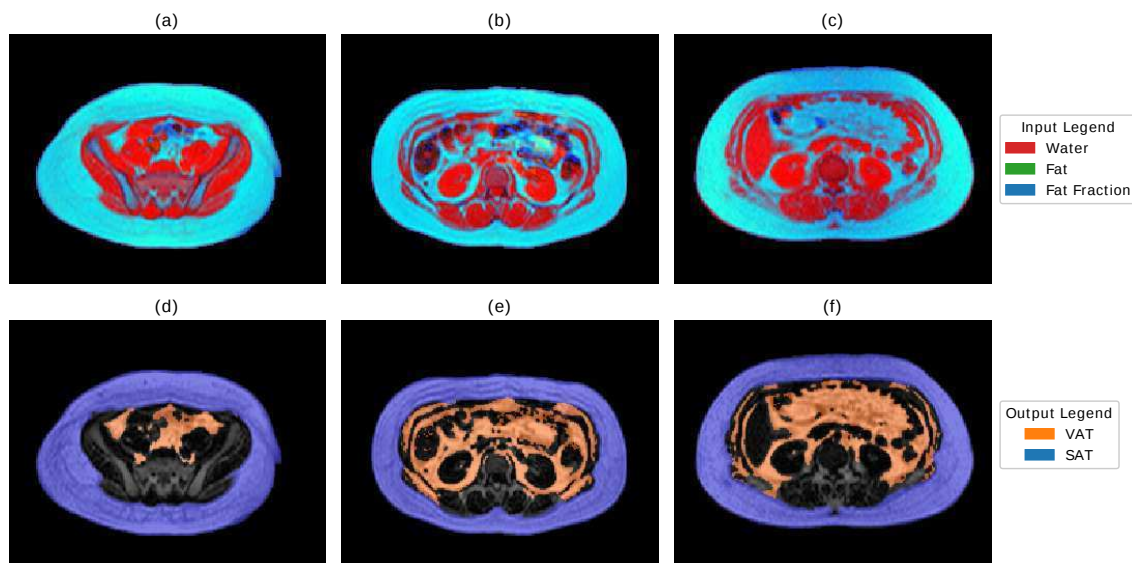}
		
		\caption{Segmentation results by the U-Net on the test dataset for the best (a, d), median (b, e) and worst (c, f) slice as measured by the sum of mislabeled voxels. The multi-channel input to the network is seen in (a-c), while (d-f) shows the respective output segmentations underneath, superimposed on the fat signal image.}
		
		\label{fig6} 
		
	\end{figure}

	Figure \ref{fig5} shows the relationship between reference segmentation volumes and error percentage of the output of the U-Net in cross-validation and on the test data, respectively. A sample of resulting segmentations is shown in Figure \ref{fig6}. For the U-Net, the most commonly observed errors in cross-validation consist of oversegmentation of the hip bones as VAT, mistakes in ambiguous areas between VAT and SAT and leakage of VAT in rare cases where skin folds lead to multiple separate SAT compartments in more obese patients. On the test data, the majority of errors occur in ambiguous areas and occasional, insular oversegmentations. When assigning VAT labels there is also a tendency towards undersegmenting some of the outer areas of the depots. For the V-Net, the used implementation is outperformed by the U-Net in all cases with the exception of a marginally higher dice score for a single SAT volume. \\
	
	When examining those image volumes that were segmented by both operators, there is similar disagreement in manual segmentation around the spine and hip bones in VAT and intramuscular tissue in SAT. The average dice scores for the overlap between these segmentations are 0.969 (VAT) and 0.975 (SAT). It is important to note that the images used for this evaluation are not the same as the test set that was used to evaluate the network, so that the dice scores can not be directly compared.	
		
	\section*{Discussion}
	
	On the given water-fat MRI data, the fully convolutional network using the U-Net architecture was able to automatically generate highly accurate VAT and SAT segmentations and clearly outperforms the V-Net. When applied to the multicenter test data, the network was robust enough for the performance to only take a minor hit despite being faced with a different patient demographic and scanning device.
	In preliminary experiments, different augmentation strategies were evaluated such as varied translations, rotations, scales and volumetric deformations. Surprisingly, none of these strategies improved the results in the final configuration, so that all networks achieved their best performance with a comparably low number of training images. The networks furthermore benefited from the hand-crafted pre-processing strategy in which the contrast was adjusted, even though they could have learned to model this step internally. The fact that the networks did not learn to perform a comparably effective pre-processing step automatically shows that the optimization process during training is not guaranteed to use the full potential of the architecture. Human decisions on the formatting of the input are therefore still a relevant factor for the network performance. \\
	
	For the U-Net, most resulting segmentations are of high quality and require no further corrections. The network is robust in regards to the background noise in the fat fraction values, even though no such noise was present in the training data. However, it was found that on average the performance on the data from Salzburg is slightly lower than on the data from the center in Uppsala, which is more similar to the training data. As a general rule the dice scores for more obese patients with larger volumes of VAT and SAT are higher than those for thin patients, a pattern that is also visible in Figure \ref{fig5}. This is probably due to the stronger effect of mislabeled single voxels in ambiguous regions in between and on the outline of the two depots. Preliminary experiments showed that modifications of the U-Net usually also achieved higher scores than the implementation of the V-Net, which indicates that the volumetric architecture may not be well suited for the given data. However, the underlying reasons for this effect can not easily be determined without further extensive experiments which are beyond the scope of this work. In our comparison the U-Net is more flexible, trains faster and generalizes better than the V-Net on the given scans. \\
	
	Both networks perform worse on the test data than in cross-validation and have a tendency to undersegment both VAT and SAT. It is likely that such a decrease in performance is not only due to the different study population and the multicenter data in the test dataset. As observed in the representative sample of training images that were segmented by both operators, the average dice scores for the inter-operator variability are in the same range or slightly lower than the network performance on the test data. Similar values have been previously reported for the agreement between manual segmentations by different operators \cite{joshi2013automatic}. Even though the values presented here were obtained on different image sets, so that they do not allow for a direct comparison between the network and human performance, they indicate that the potential for further significant improvements is largely limited by the quality of the reference segmentations. 
	Despite these observations, the lower performance on the test set as compared to the cross-validation indicates that the presented method should not be understood as a general solution to VAT and SAT segmentation in water-fat MRI. The approach can not be expected to retain this level of performance when applied to more strongly deviant images that, for example, strongly differ in position, image contrast or the usage of surface coils.

	\section*{Conclusion}
	
	In conclusion, the experiments show that the proposed strategy is able to generate accurate and robust automated VAT and SAT segmentations in water-fat MRI scans with the help of the U-Net architecture. The segmentations allow for a volume quantification with an average of absolute errors below one percent in cross-validation and below three percent when applied to the multicenter test data. When examining the dice scores, the performance of the network on the test data is within range of the observed variability between different human operators both reported here and in the literature. In a direct comparison of network architectures, the implementation of the volumetric V-Net was clearly outperformed and less robust than the U-Net. \\	
	The results of the cross-validation suggest that, when given a comparable number of (N = 90) already segmented images as training data, this approach could be employed in the context of a large-scale study to automatically segment the remaining scans. The high degree of robustness of the approach in regards to differences in patient demographics as well as the usage of a different device (of the same vendor) is seen in the results on the multicenter test data. This result shows the potential of the strategy to successfully process data of studies without any existing segmentations. In practice, the network could accordingly be trained using existing reference data and be applied to future studies using the same imaging protocol in order to provide automated segmentations with a minimal need for manual intervention.

	\section*{Acknowledgments}
	
	The research leading to these results has received funding from the European Union's Seventh Framework Program (FP7/2007-2013) under grant agreement number 279153.

	
	\bibliographystyle{unsrt}

	\bibliography{library}
	
	\pagebreak

\end{document}